\documentclass[12pt,a4paper]{article}
\usepackage[utf8]{inputenc}
\usepackage{amsmath}
\usepackage{amsfonts}
\usepackage{amssymb}
\usepackage{changepage}
\usepackage[english]{babel}

\usepackage{natbib}
\usepackage{graphicx}
\setcitestyle{authoryear}
\usepackage{float}
\usepackage[margin=0.5in]{geometry}
\usepackage[ruled,linesnumbered]{algorithm2e}
 
\usepackage{bm}

\SetArgSty{textnormal}
\newcommand{\x}{\text{x}}
\newcommand{\y}{\text{y}}
\newcommand{\dd}{\text{d}}
\newcommand{\bc}{\text{c}}
\newcommand{\s}{\text{s}}
\newcommand{\abar}{\bar{\alpha}}
\newcommand{\abart}{\bar{\alpha_{t}}}
\newcommand{\pindent}{\hspace{\parindent}}
\usepackage{hyperref}

\graphicspath{ {./} }

\begin{document}
\begin{center}
\begin{huge}
Diffusion models for Handwriting Generation\\
\end{huge}
\end{center}

\vspace{0.25cm}

\begin{center}
\begin{large}

\textbf{Troy Luhman}\footnote{\label{note1}Equal contribution} \hspace{4cm} \textbf{Eric Luhman}\footnotemark[1] \\
troyluhman@gmail.com \hspace{4cm} ericluhman2@gmail.com
\end{large}
\end{center}
\vspace{0.25cm}



\begin{center}
\textbf{Abstract}
\\

\end{center}
\begin{changemargin}{2cm}{2cm} 
In this paper, we propose a diffusion probabilistic model for handwriting generation. Diffusion models are a class of generative models where samples start from Gaussian noise and are gradually denoised to produce output. Our method of handwriting generation does not require using any text-recognition based, writer-style based, or adversarial loss functions, nor does it require training of auxiliary networks. Our model is able to incorporate writer stylistic features directly from image data, eliminating the need for user interaction during sampling. Experiments reveal that our model is able to generate realistic , high quality images of handwritten text in a similar style to a given writer. Our implementation can be found at \url{https://github.com/tcl9876/Diffusion-Handwriting-Generation}.
\end{changemargin}


\section{Introduction}

Deep generative models have been able to produce realistic handwritten text. Handwriting data can be stored in online or offline format. Online data is rendered as a sequence of pen strokes, and offline data is stored directly as an image. Online temporal data is easier to work with due to its lower dimensionality, but is harder to collect since each stroke must be recorded during writing. Each sample of handwriting data is associated with a text sequence label describing the content of the handwritten text. Each handwritten sample can also be associated with another example of handwriting from the same writer, which provides information about the style of the writer. 
\\

The task of handwriting generation was first done by \citep{graves}, who used a RNN to synthesize online handwriting. Since then, most work on online handwriting generation has used an RNN based architecture, while most work on offline handwriting generation has used a GAN. Both methods have been able to generate realistic, diverse samples in a given writer’s handwriting style. RNN based models \citep{handgen_ar1, handgen_ar2, handgen_ar3} are simple to train and sample from, but one challenge they face is that they require the writer style data to be in online format. This poses a challenge for incorporating stylistic information, as the user’s pen strokes must be recorded during sampling. GAN based methods for handwriting generation \citep{handgen_gan1, handgen_gan2, handgen_gan3} have been used as an alternative to online autoregressive models. GAN based methods have the advantage of being able to incorporate offline stylistic information, as they work with images directly. The downside to GAN-based methods is that they can be difficult to train and may suffer from lack of diversity due to mode collapse. Additionally, previous work on GAN-based handwriting generation requires training at least two auxiliary networks: the discriminator and a text recognition network to control the content of the generated text. 
\\

Diffusion probabilistic models \citep{diffusionmodels, diffusionimage} use a Markov chain to convert a known distribution (e.g. Gaussian) into a more complex data distribution. A diffusion process converts the data distribution into a simple distribution by iteratively adding Gaussian noise to the data, and the generative model learns to reverse this diffusion process. Diffusion models can be trained to optimize a weighted variational lower bound of the data likelihood. This objective is similar to that of noise conditional score networks \citep{songermon}, which estimates the gradients of the data distribution. Both noise conditional score networks and diffusion models generate samples by starting from Gaussian noise and gradually removing this noise.
\\

In this paper, we propose a diffusion probabilistic model for online handwriting generation. Our proposed model has advantages over both autoregressive and GAN based methods of handwriting generation. Our model, though it generates samples in online format, is able to incorporate writer stylistic features from offline data, eliminating the need for user interaction during sampling. Our method of handwriting generation does not require using any text based, style based, or adversarial loss functions, nor does it require training of auxiliary networks. As a result, our model has a very simple training and sampling procedure. Our model is able to generate realistic samples of handwritten text in a similar style to the original writer.

\section{Diffusion Models for Handwriting Generation}

\subsection{Diffusion Probabilistic Models}

Let $q(\y_0)$ be the data distribution, and let $\y_1$, ..., $\y_T$ be a series of $T$ latent variables with the same dimensionality as $\y_0$. A diffusion model consists of two processes: a noise adding diffusion process, and a reverse process \citep{diffusionmodels}. The posterior $q\left(\y_{1:T}\ \mid \y_{0}\right)$, or the diffusion process, is defined as a fixed Markov chain where Gaussian noise is added at each iteration based on a fixed noise schedule $\beta_1, \ldots, \beta_T$:
\begin{align} \label{eq:1}
q\left(\y_{1:T}\ \mid \y_{0}\right)=\ \prod_{t=1}^{T}q\left(\y_{t\ } \mid \y_{t-1}\right) , \pindent
q\left(\y_{t}\ \mid \y_{t-1}\right)=\ \mathcal{N}\left(\y_{t} ; \sqrt{1-\beta_{t}}\y_{t-1},\beta_{t}\bm{I}\right)
\end{align}

The reverse process is defined as a Markov chain parameterized by $\theta$:
\begin{align} \label{eq:2}
p(\y_{T}) = \mathcal{N}(\y_T; 0, \bm{I}), \pindent
p_{\theta}\left(\y_{0:T}\ \right)=p\left(\y_{T}\right)\ \prod_{t=1}^{T}p_{\theta}\left(\y_{t\ }\mid \y_{t-1}\right)
\end{align}
where $p_{\theta}\left(\y_{t-1}\ \mid \y_{t}\right)$ intends to reverse the effect of the noise adding process $q\left(\y_{t}\ \mid \y_{t-1}\right)$:
\begin{align} \label{eq:3}
p_{\theta}\left(\y_{t-1}\ \mid \y_{t}\right)=\ \mathcal{N}\left(\y_{t-1};\mu_{\theta}\left(\y_{t},t\right),\ \Sigma_{\theta}\left(\y_{t},t\right)\right)
\end{align}
where $\Sigma_{\theta}\left(\y_{t},t\right)=\sigma_{t}^{2}\bm{I}$ and $\sigma_{t}^{2}$ is a constant related to $\beta_t$. The forward process posterior is defined as:
\begin{align} \label{eq:4}
q(\y_{t-1} \mid \y_t, \y_0) = \mathcal{N}(\y_{t-1} ; \tilde{\mu}\left(\y_{t},\y_0\right), \sigma_{t}^{2}\bm{I})
\end{align}

\citep{diffusionimage} showed that the ELBO can be calculated in closed form by expanding it into a series of KL Divergences between Gaussian distributions:
\begin{align} \label{eq:5}
\textbf{ELBO} = - \mathbb{E}_{q}\left(D_{KL}\left(q(\y_T \mid \y_0) \ \middle\| \ p(\y_T) \right)+\sum_{t=2}^{T}D_{KL}\left(q(\y_{t-1} \mid \y_t, \y_0) \ \middle\| \ p_{\theta}(\y_{t-1} \mid \y_t)\right)-\log p_{\theta}\left(\y_{0} \mid \y_{1}\right)\right)
\end{align}

By defining some constants, $\y_t$ can be calculated in closed form for any step \textit{t}:
\begin{align} \label{eq:6}
\alpha_t = 1-\beta_t, \ \bar{\alpha_t} = \prod_{s=1}^{t}\alpha_{s}, \ \epsilon \thicksim \mathcal{N}(0, \bm{I}), \ \y_t = \sqrt{\abart}\y_{0}+\sqrt{1-\abart}\epsilon
\end{align}

The loss function then becomes the following for some timestep $t-1$, where $\mu_{\theta}$ is a model that predicts the forward process posterior mean $\tilde{\mu}$:
\begin{align} \label{eq:7}
L_{t-1} = E_{\y_0,\epsilon}\left(\frac{1}{2\sigma^{2}}\left\|\frac{1}{\sqrt{\alpha_t}}\left(\y_{t}-\frac{\beta_t}{\sqrt{1-\abart}}\epsilon\right)-\mu_{\theta}\left(\y_{t}, t\right)\right\|^{2}\right)
\end{align}

\citep{diffusionimage} observed that diffusion probabilistic models can be reparameterized to resemble score-based generative models \citep{songermon}. Score-based generative models estimate the gradient of the logarithmic data density $\nabla_{x}\log\left(p\left(x\right)\right)$ with a denoising objective. Under this reparameterization, the diffusion model, $\epsilon_\theta$, predicts $\epsilon$ similar to the objective of score-based models. The loss function for predicting $\epsilon$ becomes:
\begin{align} \label{eq:8}
L_{t-1}=\mathbb{E}_{t, \epsilon}\left[C_t\|\epsilon - \epsilon_{\theta}(\y_t,t) \|_2^2\right],  C_t = \frac{\beta^{2}}{2\sigma_{t}^{2}\alpha_{t}\left(1-\bar{\alpha_t}\right)}
\end{align}

During sampling, diffusion probablilistic models iteratively remove the noise added in the diffusion process, by sampling $\y_{t-1}$ for $t = T, \ldots, 1$:
\begin{align} \label{eq:9}
\y_{t-1}=\frac{1}{\sqrt{a_{t}}}\left(\y_{t}-\frac{\beta_{t}}{\sqrt{1-\abart}}\epsilon_{\theta}\left(\y_{t} , t\right)\right) + \sigma_t z
\end{align}
where $z \thicksim \mathcal{N}(0, \bm{I})$ and $\sigma_t$ is a constant related to $\beta_t$. For our experiments, we used $\sigma_t^2 = \beta_t$.

In our experiments, we found it beneficial to make a modification to the original sampling procedure. First, as our model predicts $\epsilon$, we can use the estimate of $\epsilon$ to approximate $\y_0$ at any timestep $t$:
\begin{align} \label{eq:10}
\y_0 \approx \hat{\y}_0 = \frac{1}{\sqrt{\abart}}\left(\y_{t}-\sqrt{1-\abart}\epsilon_{\theta}(\y_t, t)\right)
\end{align}
We now use our approximation for $\y_0$ to find $\y_{t-1}$:
\begin{align} \label{eq:11}
\y_{t-1} = \sqrt{\abar_{t-1}}\hat{\y}_0 + \sqrt{1-\abar_{t-1}}z, \ z \thicksim \mathcal{N}(0, \bm{I})
\end{align}
We first estimate $\hat{\y}_0$ from $\y_t \text{and} \  \epsilon_\theta$ using Equation \ref{eq:10}, then find $\y_{t-1}$ according to Equation \ref{eq:11}. Combining these two, we get:
\begin{align} \label{eq:12}
\y_{t-1} = \frac{1}{\sqrt{\alpha_t}}\left(\y_{t}-\sqrt{1-\abart}\epsilon_{\theta}(\y_{t} , t)\right)+\sqrt{1-\abar_{t-1}}z, \ z \thicksim \mathcal{N}(0, \bm{I})
\end{align}
The training procedure is performed by minimizing Equation \ref{eq:8} with the $C_t$ term removed. We compared sampling $\y_{t-1}$ according to Equation \ref{eq:12} as opposed to Equation \ref{eq:9} and found that the Equation \ref{eq:12} led to more realistic samples (see ablation study in Table \ref{tab:1}), at the cost of a slight decrease in diversity.

\subsection{Conditional Handwriting Generation}
\begin{figure}[H] \label{fig:1}

\includegraphics[width=15cm, height=8cm]{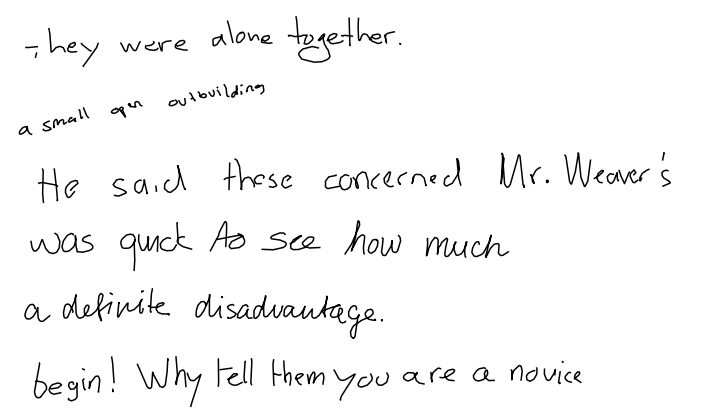} 
\caption{Samples from the IAM Database.}
\end{figure}
\pindent We will now address the task of conditional online handwriting generation. Each data point $\x_0$ is composed of a sequence of $N$ vectors $x_1 \ldots x_N$. Each individual vector in the sequence $x_n \in \mathbb{R}^2 \times \lbrace 0, 1 \rbrace$ is composed of a real valued pair which represents the pen offset from the previous stroke in the x and y direction, and a binary entry that has a value of 0 if the pen was down when writing the stroke and 1 otherwise. Each handwritten sequence is associated with a discrete character sequence c describing what was written. Each sequence is also associated with an offline image containing writer's style information, denoted by s.
\\

There is one technical issue that needs to be addressed. The reverse process in Equation \ref{eq:3} is parameterized by a Gaussian distribution. We cannot parameterize the binary variable representing whether the stroke was drawn by a Gaussian distribution as we did for the real valued pen strokes. However, we can instead parameterize it with a Bernoulli distribution, which can also be optimized in closed form.
\\

We therefore split each data point $\x_0$ into two sequences $\y_0$ and $\dd_0$ of equal length, with $\y_0$ representing the real valued pen strokes, and $\dd_0$ representing whether the stroke was drawn. At each step t, our model $\dd_\theta(\y_{t} , \bc , \s, \sqrt{\abar})$ returns an estimate $\hat{\dd_0}$ of whether the pen was down. $\dd_\theta$ shares all parameters with $\epsilon_\theta$.  Our stroke loss and pen-draw loss are shown in Equations \ref{eq:13} and \ref{eq:14}:

\begin{align} \label{eq:13}
L_{\text{stroke}}(\theta) = \|\epsilon - \epsilon_{\theta}\left(\y_t, \bc, \s, \sqrt{\abar}\right) \|_2^2
\end{align}
\begin{align} \label{eq:14}
L_{\text{drawn}}(\theta) = - \dd_{0}\log\left(\hat{\dd_0}\right)-\left(1-\dd_{0}\right)\log\left(1-\hat{\dd_0}\right)
\end{align}

We found it beneficial to weight the pen-draw loss according to the noise level, since it is more difficult to predict the pen-draws at higher noise levels. We weight Equation \ref{eq:14} by $\abar$ during training (see Algorithm \ref{algo:train}).
\\

\citep{songermon, improvedtechniques} noted that the choice of noise schedule is crucial to generating high quality samples. \citep{speechgen2} proposed to condition the model on the continuous noise level $\sqrt{\abar}$ as opposed to the discrete index $t$. This allows for the use of different noise schedules during sampling without retraining the model. In order to condition the network on the continuous noise level, we first define noise schedule \textit{l} where $l_0 = 1, l_t = \sqrt{\abart}$ During training, we condition the model on $\sqrt{\abar} \thicksim \text{Uniform}(l_{t-1}, l_t)$, where $t \thicksim \text{Uniform}(\lbrace 1, \ldots, T\rbrace)$. 
\\

One desirable property of handwriting synthesis networks is the ability to generate handwriting in the style of a given writer. Previous handwriting synthesis methods are able to control the style of the generated handwriting by conditioning on a sample written by the writer. Although our model generates online output samples, our model accepts offline images as input for writer stylistic features. Previous RNN-based models require online data during sampling, which is difficult to collect. To incorporate style information, we extract features from the image using Mobilenet \citep{mobilenet} and use these features as input to our model.
\\

We describe our training procedure in Algorithm \ref{algo:train} and our sampling procedure in Algorithm \ref{algo:test}. Note that our training procedure does not require training any auxiliary networks and does not use text recognition or style-based losses.
\\

\begin{minipage}{0.42\textwidth}
\begin{algorithm}[H]

\While{not converged} {
  $\y_0 \thicksim q(\y_0)$ \\
  $t \thicksim \text{Uniform}(\lbrace 1, \ldots, T\rbrace)$ \\
  $\sqrt{\abar} \thicksim \text{Uniform}(l_{t-1}, l_t)$ \\
  $\epsilon \thicksim \mathcal{N}(0, \bm{I})$ \\
  $\y_t = \sqrt{\abar}\y_0 + \sqrt{1-\abar}\epsilon $  \\
  Take gradient descent step on: \\
  $\nabla_\theta ( L_{\text{stroke}}(\theta) + \abar L_{\text{drawn}}(\theta) )$
}  

\caption{Training}
\label{algo:train}
\end{algorithm}
\end{minipage}
\hfill
\begin{minipage}{0.54\textwidth}
\begin{algorithm}[H]
$\y_T \thicksim \mathcal{N}(0, \bm{I})$ \\
\For{$t = T, \ldots, 1$} {
 $z \thicksim \mathcal{N}(0, \bm{I})$ \\
 $\y_{t-1} = \frac{1}{\sqrt{\alpha_t}}\left(\y_{t}-\sqrt{1-\abart}\epsilon_{\theta}(\y_{t} , \bc , \s, \sqrt{\abart})\right)$ \\
 $\y_{t-1}=\y_{t-1} + \sqrt{1-\abar_{t-1}}z$  \ if $t>1$ \\
 $\dd_0 = \dd_\theta(\y_{t} , \bc , \s, \sqrt{\abart})$
} 
\textbf{return} $\y_0, \dd_0$

\caption{Sampling}
\label{algo:test}
\end{algorithm}
\end{minipage}


\section{Model Architecture}

\begin{figure}[H] \label{fig:2} 
\includegraphics[width=14cm, height=6cm]{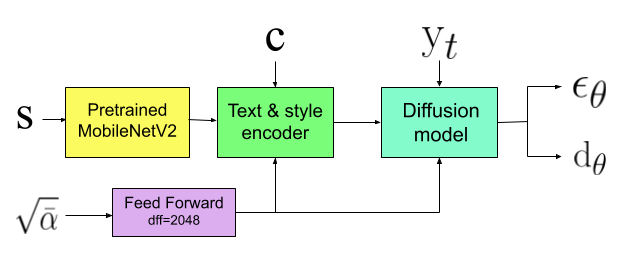} 
\caption{Overview of our full model.}
\end{figure}

Our full model is shown in Figure 2. It consists of two parts: a text and style encoder to represent the desired text and stylistic features, and a diffusion probabilistic model to predict $\epsilon_\theta$. The noise level $\sqrt{\abar}$ is passed through a feedforward network consisting of two fully connected layers. Both the encoder and the diffusion model are conditioned on the noise level.
\\

\textbf{Conditioning on the noise level:}
To condition our model on the noise level, we use affine transformations along the channel axis. Each affine transformation’s scales and biases are parameterized by the output of a fully connected layer. Our affine transformation is detailed in Figure 6 in Appendix \ref{fig:6}.
\\

\textbf{Text and Style Conditioning:}
For writer conditioning, we first extract local features from an image of their handwriting using a MobileNetV2 \citep{mobilenet} pretrained on Imagenet. Character-level embeddings are used to represent the text sequence. We then compute attention between the text sequence and the extracted features, allowing different text characters to attend to different portions of the given writer sample. This output is then added to the text sequence representation, before being passed through a feedforward network. Full architectural details on the encoder can be seen in Figure 10 in Appendix \ref{fig:10}.
\\

\textbf{Diffusion model:}
Our diffusion model consists of downsampling blocks, followed by upsampling blocks, and uses long range convolutional skip connections. We use two main types of blocks, convolutional blocks and attentional blocks. Full information on our diffusion model architecture is detailed in Figure 9 in Appendix \ref{fig:9}.

Our convolutional blocks consist of 3 convolutional layers and a convolutional skip connection. We apply conditional affine transformations to the output of every convolutional layer. Further architectural information on our convolutional block is shown in Figure 7 in Appendix \ref{fig:7}.

Our attentional blocks consist of 2 multi-head attention layers, and a feed forward network. The first attention layer performs attention between the stroke sequences x and the output of the text-style encoder, while the second performs self-attention. Sinusoidal positional encodings \citep{transformer} are added to the queries and keys at every attentional layer. For attentional layers at higher stroke resolutions, the positions of the text sequence are multiplied before the positional encoding to allow for easier alignment between the text sequence and the longer stroke sequence. We use layer normalization \citep{layernorm} followed by conditional affine transformations after every attentional layer and feed forward network. Further architectural information on our attentional block is shown in Figure 8 in appendix \ref{fig:8}.


\section{Related Work}

\pindent Our work builds off previous work on diffusion probabilistic models \citep{diffusionmodels, diffusionimage} and Noise Conditional Score Networks or NSCNs \citep{scoredenoise, songermon}, which are closely related. \citep{diffusionimage} addressed the task of image generation using a diffusion model, and \citep{speechgen1, speechgen2} used diffusion models for speech generation. \citep{songermon, improvedtechniques} used NSCNs for image generation, and \citep{shapegen} used NCSNs for shape generation. \citep{improvedsampling} combined the NCSN objective with an adversarial objective to address image generation. 
\\

Handwriting synthesis was first explored in \citep{graves}, which used a recurrent neural network (RNN) to predict each stroke one at a time. Since then, most work on online handwriting generation has made use of a type of RNN knows as a Variational RNN \citep{handgen_ar1}. Variational RNNs (VRNNs) incorporate a Variational Autoencoder \citep{vae}, or VAE , at each timestep to generate sequences. \citep{handgen_ar2} uses a conditional VRNN that is able to separate content and style elements, allowing for editing of the generated samples. \citep{handgen_ar3} incorporates character and writer level style information for further model flexibity. RNN based architectures exhibit high quality, diverse samples and have many capabilities. These drawback of autoregressive models is that they accept online writing examples in order to incorporate the writer's stylistic information, requiring the writer to interact during sampling. \citep{handgen_ar4} alleviates this problem by approximating an online representation of an offline image with the downside being that it is highly dependent on the performance of the approximation model.
\\

As an alternative to autoregressive RNN models, generative adversarial networks (GANs) \citep{gan} have been proposed for handwriting synthesize. \citep{handgen_gan1} first used a GAN to generate images of handwritten words. \citep{handgen_gan2, handgen_gan3} both use the GAN framework and improve sample quality over \citep{handgen_gan1}, in addition to being able to condition on writer stylistic features. GAN based methods of handwriting generation are not autoregressive, and are able to condition on offline style features, which is beneficial since no user interaction is required during sampling. The downside of GANs is that adversarial training procedure of GANs can be unstable and result in mode collapse, which reduces the diversity of samples produced. Previous works on handwritten text generation with GANs also have used a text recognition objective for the generated samples which can complicate the training procedure.


\section{Experiments}

\begin{figure}[H] \label{fig:3}
\includegraphics[width=20cm, height=11cm]{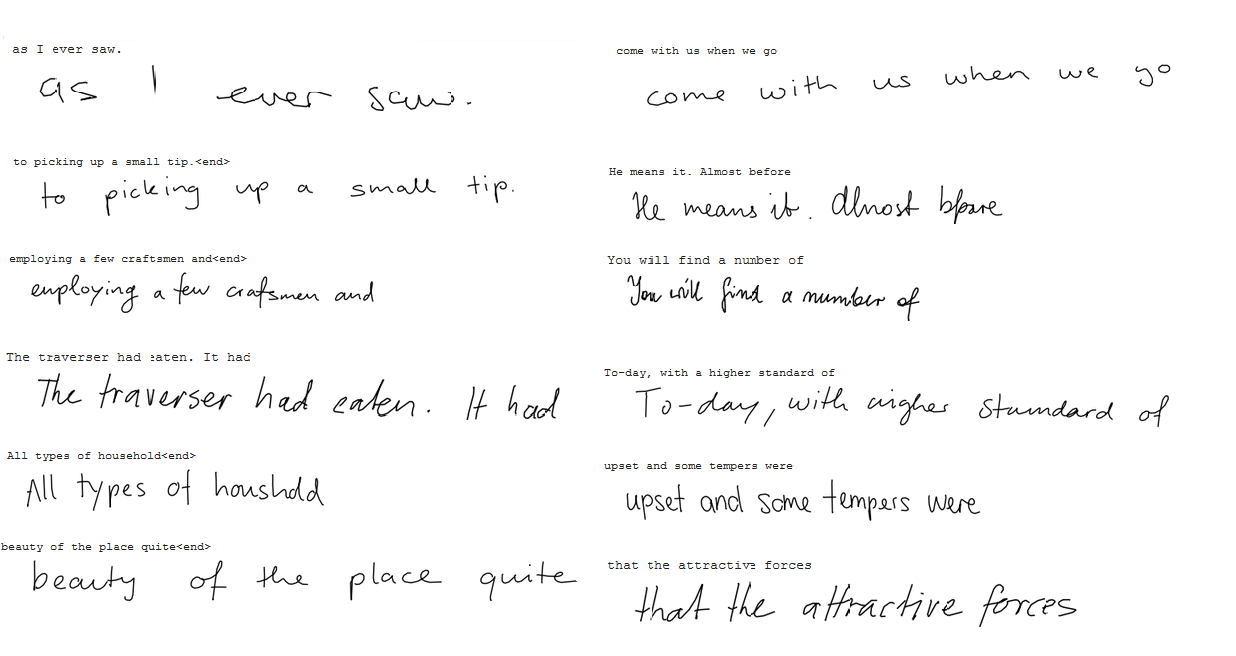} 
\caption{Four of the above samples are real, drawn from the test dataset. The remaining 8 were randomly generated by our model. We encourage the reader to try to determine which ones are fake, and which are real. Answer in Appendix \protect\ref{answers}} 

\end{figure}
\subsection{Experiment Details}

\pindent We use the IAM Online Database \citep{iamondb}, a dataset of approximately 12000 handwritten lines that are each associated with a character string label. Each sample is also associated with an sample of handwritten text by the writer, which we store as an offline image. We use the same splits as \citep{graves}. To preprocess the data, we first divide each example by its standard deviation. We discard samples containing strokes of lengths 15 standard deviations above the mean. These very long strokes usually are a result of large spacing between words. We combine strokes pointing in approximately the same direction as their neighbor, which reduces the dimensionality without visual differences.
\\

We used $T=60$ diffusion steps for our models. For our choice of noise schedule $\beta_1, \ldots, \beta_T$, we used $\beta_t$ = 0.02 + Exponential($1 \times 10^{-5}, 0.4$). Exponential($1 \times 10^{-5}, 0.4$) denotes a geometric sequence from $1 \times 10^{-5}$ to 0.4 .
We trained our model for 60000 steps with a batch size of 96 on a single Nvidia V-100.  We use the Adam optimizer \citep{adam} with $\beta_1 = 0.9$ , $\beta_2 = 0.98$, and we clip the norm of the gradients to 100. For our learning rate, we use the inverse square root schedule in \citep{transformer} with 10000 warmup steps and the $d_{model}$ argument at 256. All our models have the same size of 10.0 million parameters. These values were not sweeped over.

\pindent We evaluate the performance of our models with Frechet Inception Distance (FID) \citep{fid} and Geometry Score (GS) \citep{geometryscore}. FID uses the InceptionV3 network \citep{inception} to compare the similarity of real images to generated images. Since our model outputs online sequences as output, we first plot the sequences and then convert them to offline images. The Geometry score compares the geometric properties of the real and generated data manifolds to measure quality and diversity. 
\\

To conduct the experiment, we draw two sets of 4500 examples from the training dataset. We compare the first set of examples with our model's predictions of the second set of examples given the corresponding text and style information. For the FID evaluation, we resize all the plotted images to size M $\times$ N. For the Geometry Score evaluation, we keep the generated data in online form, and use the same values of $L_0, \gamma, i_{max}$ as \citep{geometryscore}. We run both experiments once.

\subsection{Results and Discussion}
We report our scores for the objective evaluation metrics in Table \ref{tab:1}. 
\begin{table}[H]
  \begin{center}
    \caption{Objective evaluation metrics (lower is better)}
    \label{tab:1}
    \begin{tabular}{l|c|r} 
      \ & \textbf{FID} & \textbf{GS}\\
      \hline
      Ground Truth & 2.91 & $5.4 \times 10^{-4}$ \\
      Our Model & 7.10 &  $3.3 \times 10^{-3}$\\
      Ablated Model & 8.05 &  $2.7 \times 10^{-3}$\\
    \end{tabular}
  \end{center}
\end{table}

Due to differences in the dataset and format of the generated output, we do not directly compare our results with previous works. Our images for FID are first plotted on a graph, then the graph is converted to an image. This is in contrast to previous methods that used FID, which generated offline images directly. Our GS is computed from online data, which cannot be compared with scores from offline data. We instead compare our results with online samples from the dataset itself. The ground truth score is calculated with the real examples in the second set as opposed to the model's predictions of examples in the second set. 
\\

\textbf{Ablation Study} We conduct an ablation study to compare the quality of samples generated with our sampling procedure (in Algorithm \ref{algo:test}) to those generated with the original sampling procedure, which uses Equation \ref{eq:9}. We compare FID and GS for the generated samples in Table \ref{tab:1}. We used the same trained model for both models, and only changed the sampling procedure. We found that in the case of online handwriting generation, our sampling procedure yielded relatively higher FID scores but relatively lower Geometry Score. For qualitative comparison between the two, see Figures 15 and 16 in the Appendix.
\\

\textbf{Attention Weights} Learning to properly align the text sequence to the stroke sequence is critical for generating longer sequences of realistic text without missing, misplaced or confused letters. The diagonal line in Figure 4 represents the predicted alignment between the strokes and the text sequence. Empirically, when the attention weights diverged from the diagonal line, this led to spelling mistakes. We observed that the text-to-stroke attention weights remain well-aligned throughout the reverse process, even in the earliest stages where the text itself is not recognizable. 

\begin{figure}[H] \label{fig:4}
\includegraphics[width=16cm, height=7cm]{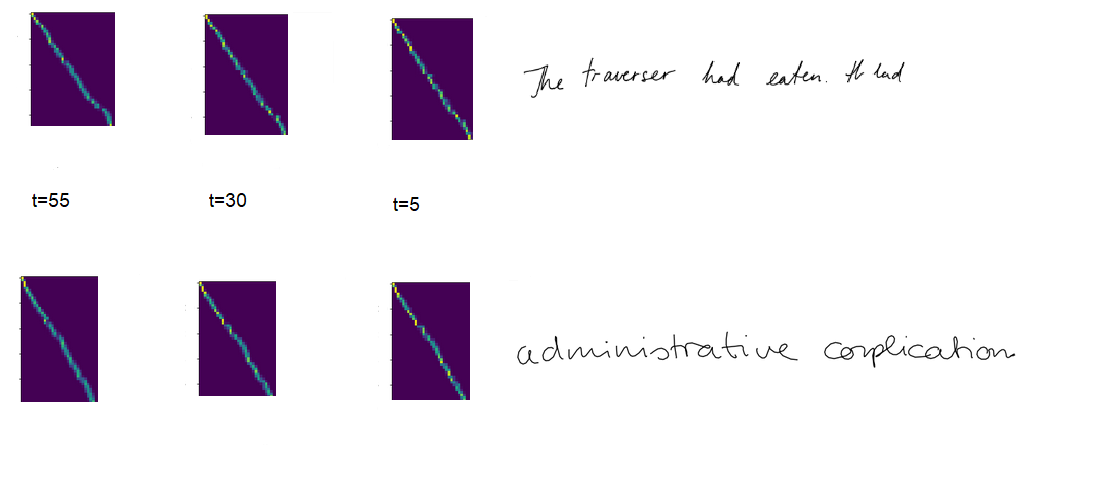} 
\caption{Text - to - stroke attention weights at $t=5, t=30, t=55$. The diagonal line shows the alignment predicted by the model between the strokes and the texts. }
\end{figure}

\textbf{Style Interpolation}
Given two samples of handwriting from two different writers, we would like  to be able to interpolate a sample that shares stylistic aspects with both writers. In order to incorporate style information, our model accepts a feature vector s that is obtained by extract features from the given image using MobileNet. To perform style interpolation, we first obtain feature vectors $\s_0', \  \s_1'$ from $\s_0, \ \s_1$ and condition our model on $\hat{s}$, where $\hat{s} = \lambda \s_0' + (1-\lambda) \s_1'$. $\lambda \in [0, 1]$ controls the relative importance of $\s_0'$.

\begin{figure}[H] \label{fig:5}
\includegraphics[width=20cm, height=11cm]{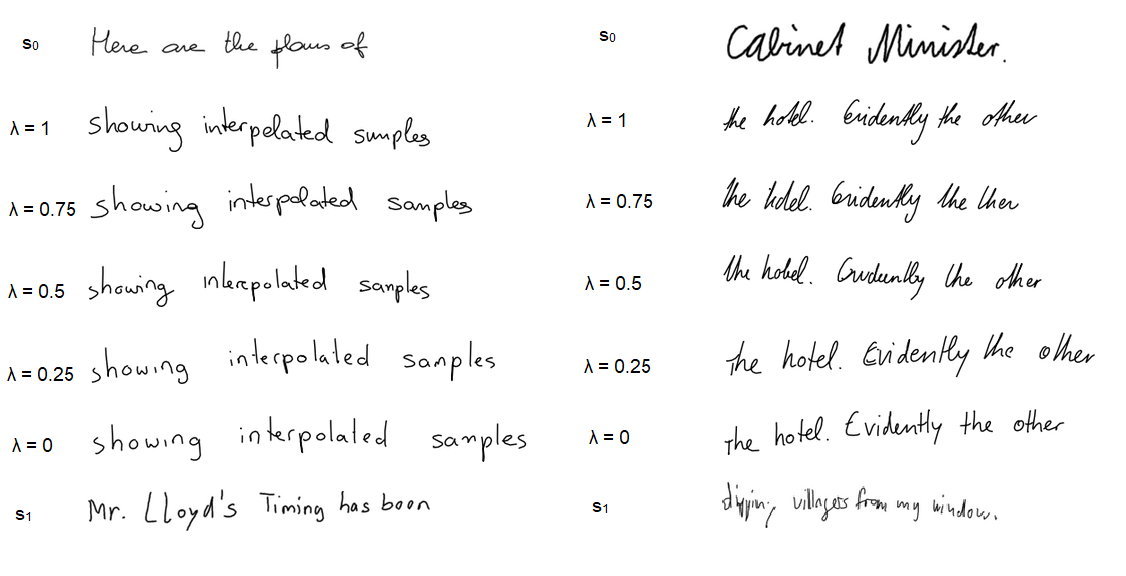} 
\caption{Style Interpolation. The top and bottom images $\s_0, \  \s_1$ contain writer style information and are drawn from the test dataset. To perform style interpolation, we keep the generated text constant, and gradually decrease $\lambda$. The generated samples share calligraphic aspects with the bottom picture $\s_1$ more and less with the top picture $\s_0$ less as $\lambda$ decreases. }
\end{figure}


\section{Conclusion}
In this paper, we have proposed a diffusion probabilistic model for online handwriting generation. Our work builds off of previous work in both diffusion probabilistic models \citep{diffusionimage}, and closely related score based generative models \citep{songermon}. Our training procedure is more simple than most previous handwriting generation methods, as ours does not require adversarial, text-recognition based or writer-recognition based objectives. We also demonstrate the ability for our model to incorporate calligraphic features from offline data, which is easier to collect during sampling time. Empirically, the samples from our model are realistic and diverse.

\nocite{*}
\bibliography{references}

\appendix
\section{Architecture Details}

\begin{minipage}{0.5\textwidth}
\begin{figure}[H] \label{fig:6}
\includegraphics[width=4cm, height=4cm]{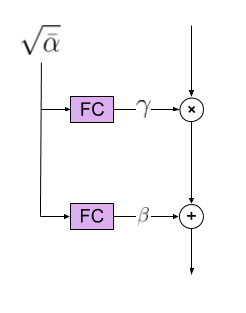} 
\caption{Our Conditional Affine Transformation.}
\end{figure}
\begin{figure}[H] \label{fig:7}
\includegraphics[width=8cm, height=10cm]{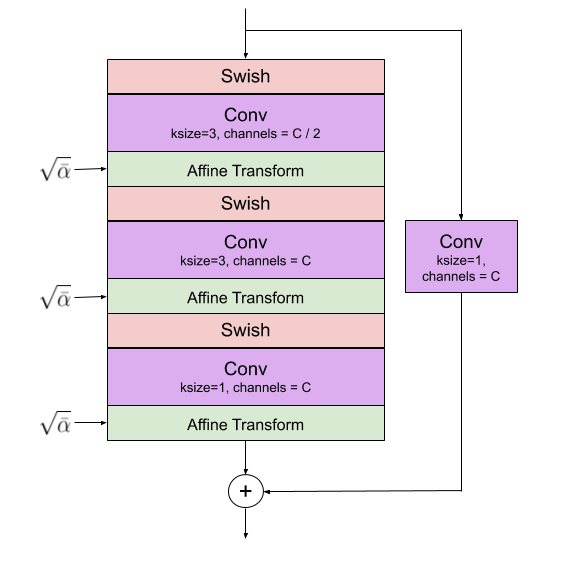} 
\caption{Our Convolutional block.}
\end{figure}
\end{minipage}
\hfill
\begin{minipage}{0.47\textwidth}
\begin{figure}[H] \label{fig:8}
\includegraphics[width=9cm, height=14cm]{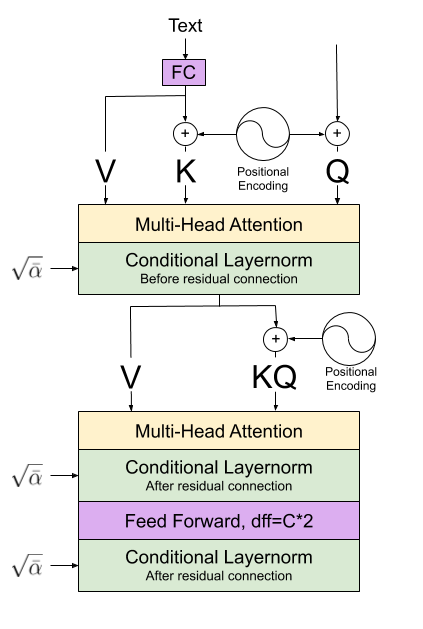} 
\caption{Our Attentional block.}
\end{figure}
\end{minipage}

\begin{minipage}{0.47\textwidth}
\begin{figure}[H] \label{fig:9}
\includegraphics[width=9cm, height=14cm]{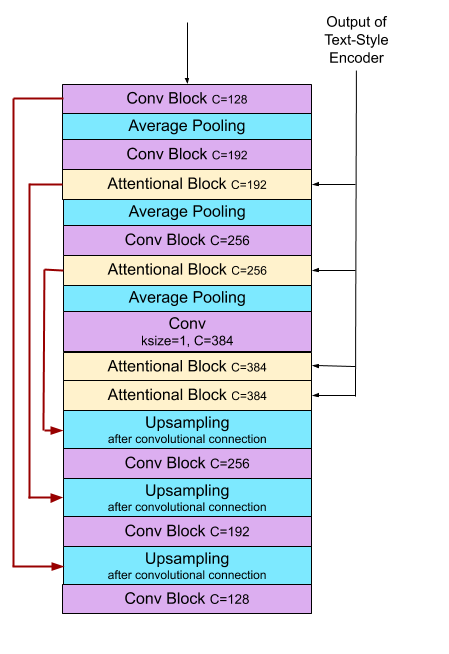} 
\caption{Architecture of the diffusion model.}
\end{figure}
\end{minipage}
\hfill
\begin{minipage}{0.47\textwidth}
\begin{figure}[H] \label{fig:10}
\includegraphics[width=11cm, height=11cm]{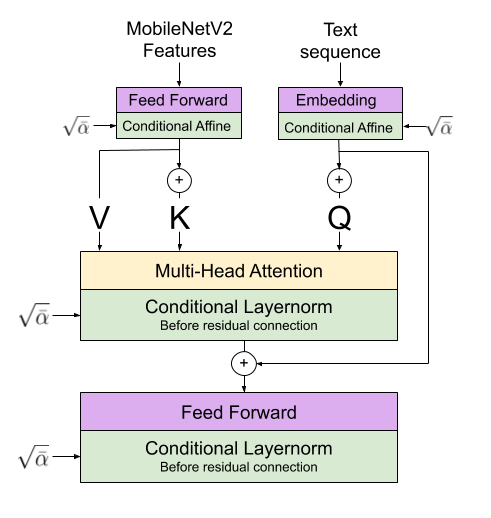} 
\caption{Architecture of the Encoder.}
\end{figure}
\end{minipage}

\section{Samples}
The phrases "as I ever saw", "that the attractive forces", "the traverser had eaten", and "you will find a number of" in Figure \ref{fig:3} are real.
\label{answers}

\begin{figure}[H] \label{fig:11}
\includegraphics[width=18cm, height=9cm]{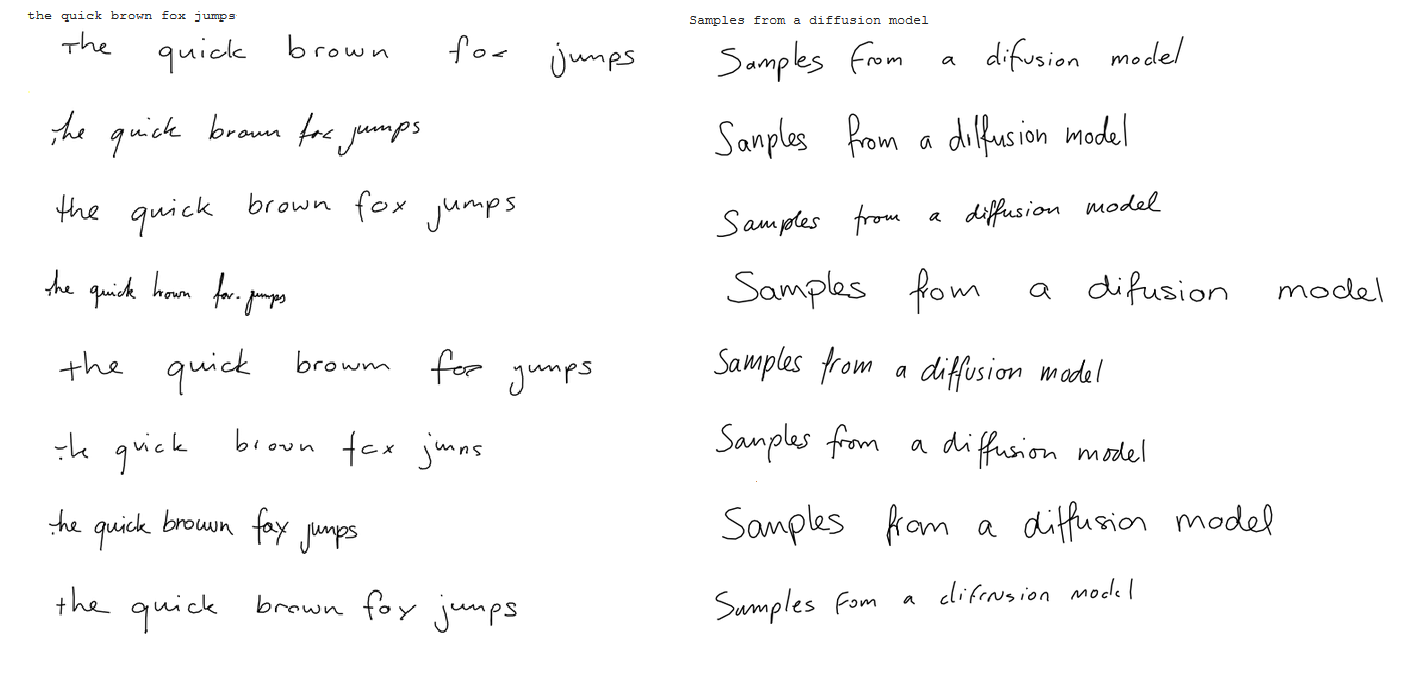} 
\caption{Samples generated from our model. On the left: "the quick brown fox jumps". On the right: "Samples from a diffusion model"}
\end{figure}

\begin{figure}[H] \label{fig:12}
\includegraphics[width=19cm, height=10cm]{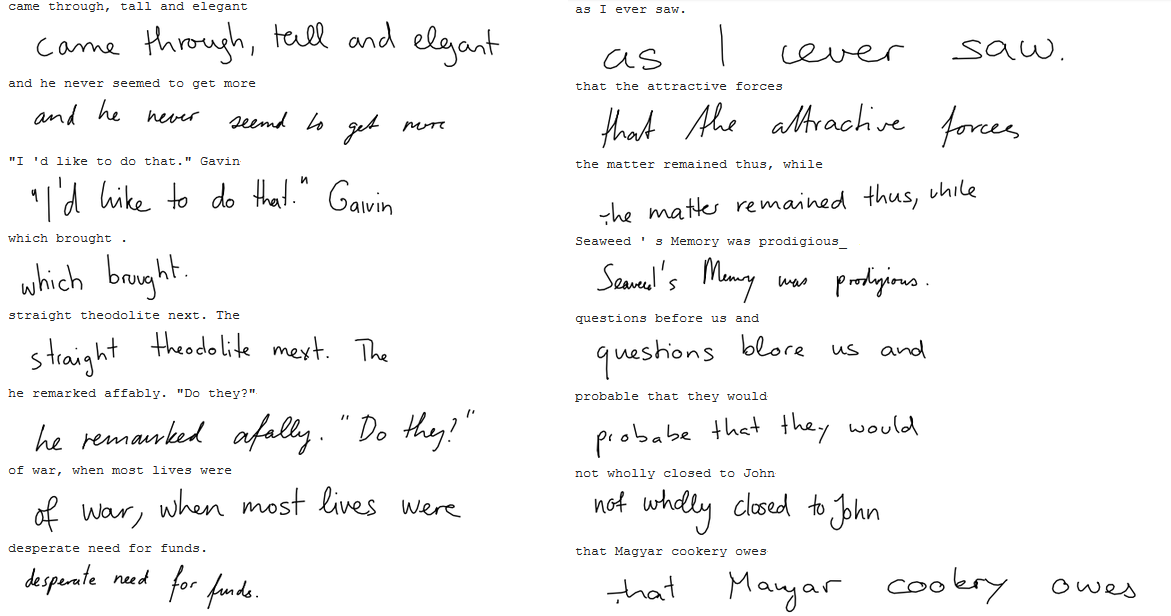} 
\caption{Samples generated from our model.}
\end{figure}

\begin{figure}[H] \label{fig:13}
\includegraphics[width=18cm, height=9cm]{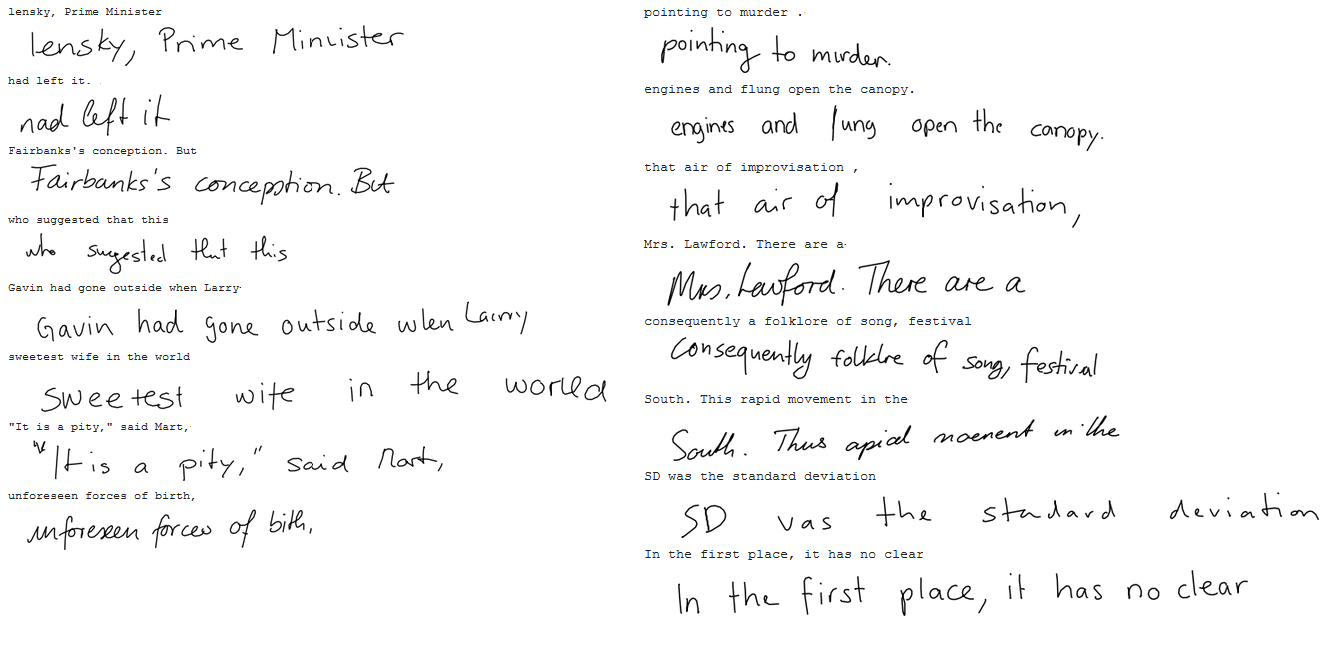} 
\caption{Samples generated from our model.}
\end{figure}

\begin{figure}[H] \label{fig:14}
\includegraphics[width=19cm, height=9cm]{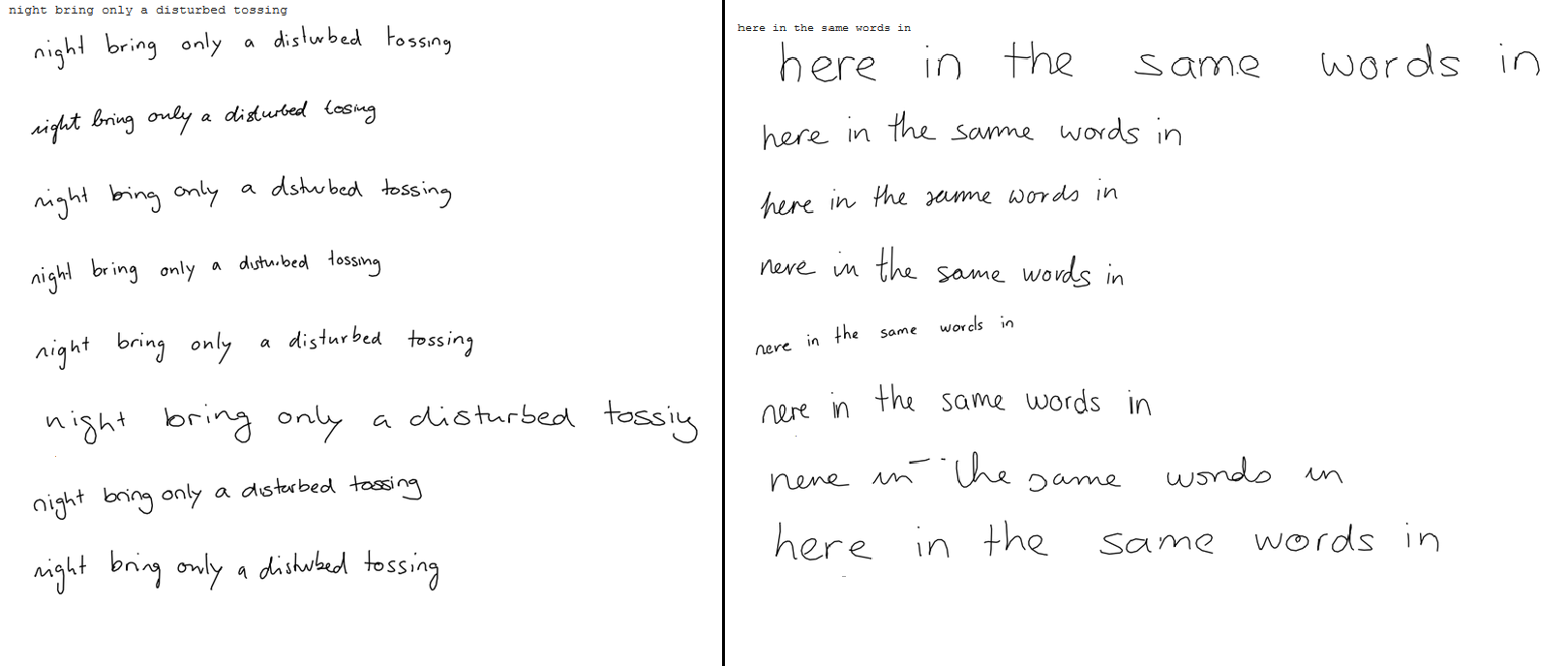} 
\caption{Samples generated from our model. Both of the texts, "night bring only a disturbed tossing" and "here in the same words in" are from the test dataset.}
\end{figure}

\begin{figure}[H] \label{fig:15}
\includegraphics[width=18cm, height=4.5cm]{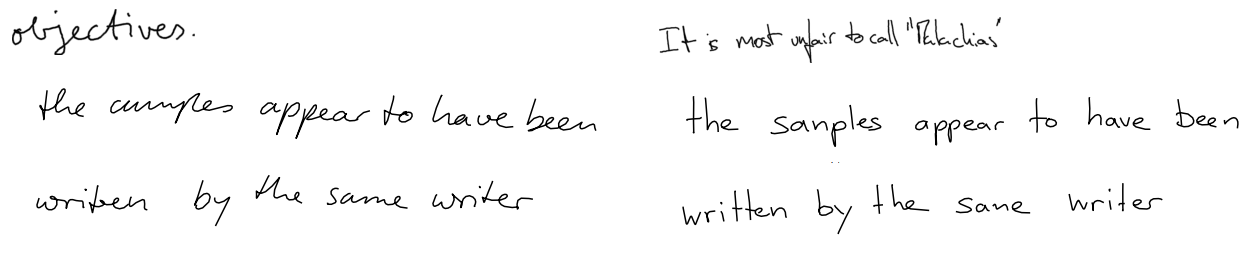} 
\caption{In each column, the top line is a sample of a writer from the test dataset's style. The bottom two lines are produced by the generative model using the style information s in the first line. }
\end{figure}

\begin{figure}[H] \label{fig:16}
\includegraphics[width=18cm, height=9cm]{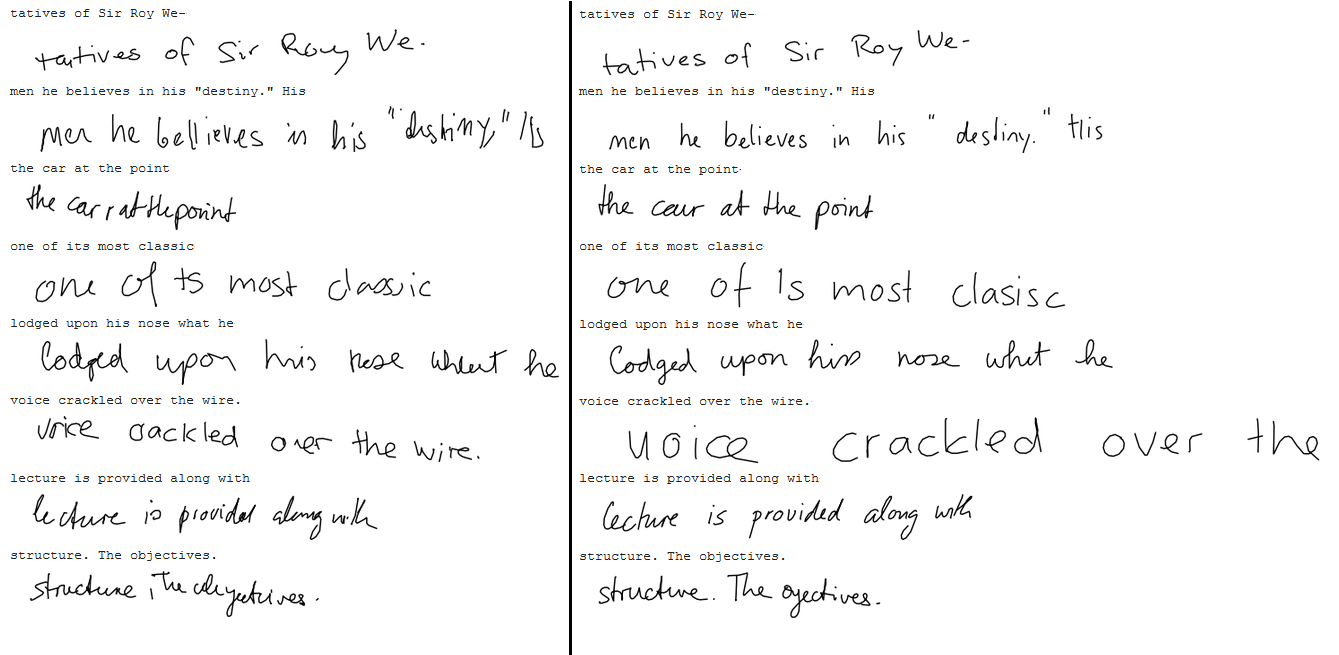} 
\caption{The left column shows samples from the \textit{ablated} model, which uses Equation \ref{eq:9} to compute $\y_{t-1}$. The right column shows samples from our final model. }
\end{figure}

\begin{figure}[H] \label{fig:17}
\includegraphics[width=18cm, height=9cm]{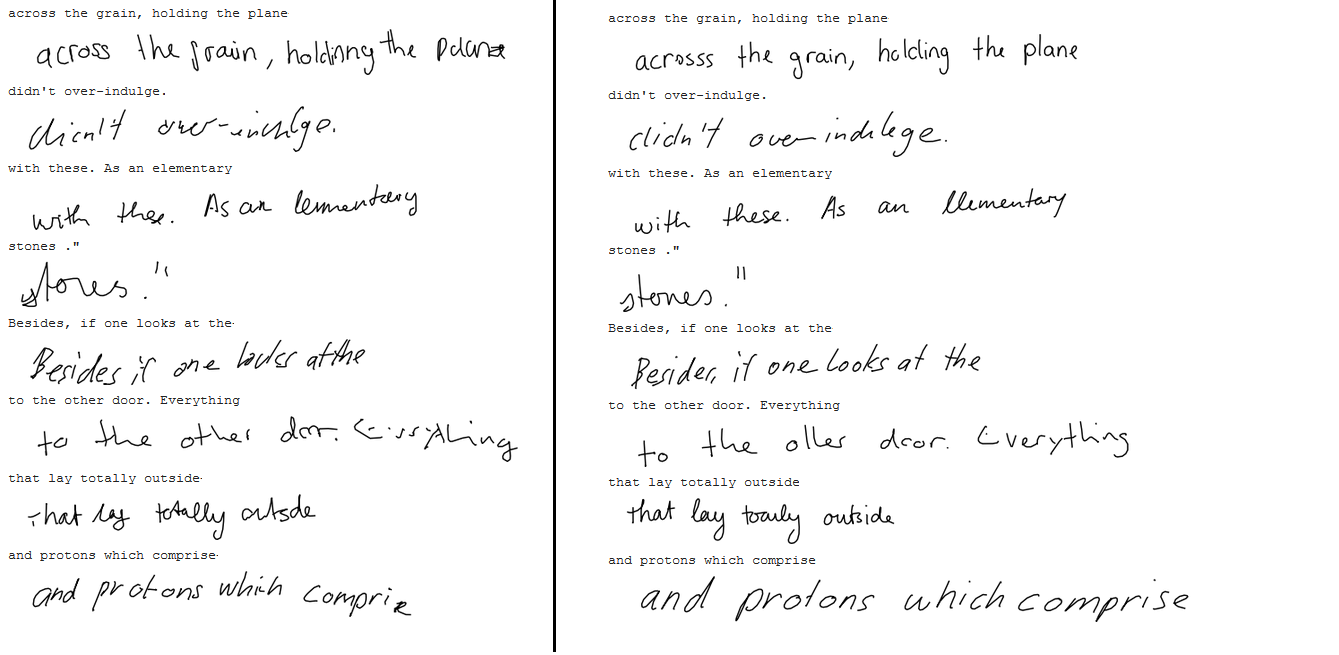} 
\caption{The left column shows samples from the \textit{ablated} model, which uses Equation \ref{eq:9} to compute $\y_{t-1}$. The right column shows samples from our final model.}
\end{figure}
\label{appb}
\end{document}